%% file: main.tex
\documentclass[10pt,twocolumn,letterpaper]{article}

\usepackage{times}
\usepackage{xspace}
\usepackage[dvipsnames]{xcolor}
\usepackage{graphicx}
\usepackage{amsmath}
\usepackage{amssymb}
\usepackage{booktabs}
\usepackage[numbers,sort&compress]{natbib}
\usepackage{silence}
\usepackage[format=plain,labelformat=simple,labelsep=period,font=small,compatibility=false]{caption}
\usepackage[font=footnotesize,skip=3pt,subrefformat=parens]{subcaption}
\usepackage[hyphens]{url}
\usepackage[shortlabels,inline]{enumitem}
\usepackage{makecell}

\setlist[itemize]{noitemsep,leftmargin=*,topsep=0em}
\setlist[enumerate]{noitemsep,leftmargin=*,topsep=0em}

\makeatletter
\DeclareRobustCommand\onedot{\futurelet\@let@token\@onedot}
\def\@onedot{\ifx\@let@token.\else.\null\fi\xspace}

\makeatletter
\def\section{\@startsection {section}{1}{\z@}
   {-10pt plus -2pt minus -2pt}{7pt} {\large\bf}}
\def\subsection{\@startsection {subsection}{2}{\z@}
   {-8pt plus -2pt minus -2pt}{5pt} {\bf}}
\def\subsubsection{\@startsection {subsubsection}{3}{\z@}
   {-6pt plus -2pt minus -2pt}{3pt} {\bf}}
\makeatother

\renewenvironment{abstract}{%
   \centerline{\large\bf Abstract}%
   \vspace*{12pt}\noindent%
   \it\ignorespaces%
}{%
   \vspace*{12pt}%
}

\input{preamble}

\usepackage[pagebackref,breaklinks,colorlinks,allcolors=blue]{hyperref}

\usepackage[capitalize]{cleveref}
\crefname{section}{Sec.}{Secs.}
\Crefname{section}{Section}{Sections}
\Crefname{table}{Table}{Tables}
\crefname{table}{Tab.}{Tabs.}

\title{OffNadirLoc: Benchmark and Framework for Challenging UAV-to-Satellite Geo-Localization under Large Off-Nadir Views}

\author{
Qian Qiao$^{1*}$ \quad
Wenye Liu$^{1*}$ \quad
Ting Liu$^{1\dagger}$ \quad
Jiuhe Shu$^{1}$ \quad
Peng Wang$^{1}$\\
$^{1}$School of Computer Science, Northwestern Polytechnical University\\
{\tt\small \{qianqiao, liuwenye, shujiuhe\}@mail.nwpu.edu.cn, \{liuting, peng.wang\}@nwpu.edu.cn }\\
}

\begin{document}
\maketitle
\footnotetext[1]{$^{*}$ Equal contribution. $^{\dagger}$ Corresponding author.}

\begin{figure*}[!h]
  \centering
  \includegraphics[width=\linewidth]{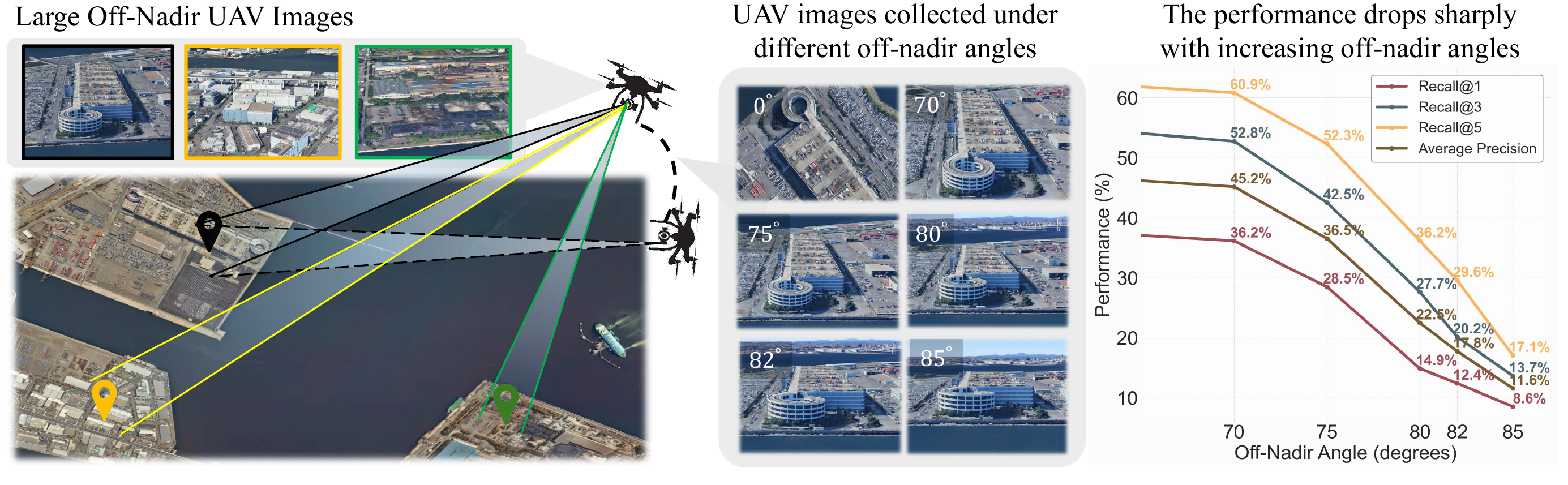}
  \vspace{-0.8cm}
  \caption{OffNadirLoc benchmark overview. UAV images captured under large off-nadir angles (70°, 75°, 80°, 82°, 85°) exhibit severe perspective distortion, occlusion, and appearance shifts relative to nadir satellite views, forming a challenging setting for cross-view geo-localization.}
  \vspace{-0.2cm}
  \label{fig:dataset}
\end{figure*}

\input{sec/0_abstract}
\input{sec/1_intro}

\input{sec/2_related_work}
\input{sec/3_dataset}
\input{sec/4_method}

\input{sec/5_experiment}
\input{sec/6_conclusion}
{
    \small
    \bibliographystyle{ieeenat_fullname}
    \bibliography{main}
}

\end{document}

%% file: preamble.tex
\usepackage{float}

\newcommand{\TODO}[1]{}



%% file: sec/0_abstract.tex
\begin{abstract}
Cross-view geo-localization between UAV and satellite imagery remains a fundamental yet highly challenging task, especially under large off-nadir views where drastic perspective distortions, occlusions, and appearance gaps occur. Existing benchmarks and methods primarily focus on near-nadir scenarios and often overlook the importance of structural scene understanding and intra-domain relational constraints, limiting their performance in real-world deployments.
In this work, we introduce \textbf{OffNadirLoc}, a new benchmark for large off-nadir UAV-to-satellite geo-localization. 
To tackle the unique challenges posed by off-nadir perspectives, we further propose ONLoc, a framework that incorporates a structure-aware contextual weighting mechanism to dynamically emphasize reliable local features while suppressing ambiguous or repetitive regions. Additionally, we design a view-coherent learning strategy, which treats one satellite image and the corresponding UAV images from multiple views as a cohesive semantic group. This set-level supervision enables the model to learn viewpoint-invariant and discriminative features, making it more effective at capturing multi-view consistency than conventional pairwise contrastive learning. Extensive experiments on the OffNadirLoc benchmark and four near-nadir datasets demonstrate that our method consistently outperforms state-of-the-art approaches while exhibiting strong zero-shot generalization to unseen datasets without additional training. The code will be released at \url{https://montalario.github.io/offnadirloc/}.
\end{abstract}

%% file: sec/1_intro.tex
\section{Introduction}

Visual geo-localization aims to determine the geographic location of an input image by matching it to geo-tagged satellite imagery~\cite{tian2021uav,berton2022rethinking,wilson2024image}. With the increasing availability of high-resolution remote sensing data, geo-localization using aerial and satellite imagery has attracted significant research interest. Many real-world scenarios require target/object localization from UAV imagery, such as urban planning, ecological monitoring, and disaster response~\cite{berton2022deep,durgam2024cross,xia2024enhancing}. 
Most existing approaches assume that the input images are captured from nadir or near-nadir viewpoints, where geometric alignment is more straightforward and visual appearance is relatively consistent. These assumptions simplify the localization process and have enabled promising results in controlled conditions.


However, in certain practical scenarios, obtaining nadir-view imagery is either infeasible or highly constrained. For instance, when UAVs operate in sea areas, near mountain cliffs, or during rapid-response missions in complex terrain, the camera may only capture the scene from a large off-nadir angle due to restrictions in flight path, safety concerns, or sensor orientation limits. As illustrated in Figure~\ref{fig:dataset}, such imagery suffers from significant perspective distortions and occlusions, resulting in geometric misalignments relative to satellite maps, which substantially complicates accurate geo-localization. Despite its practical importance, this large off-nadir setting remains underexplored.

Existing UAV-satellite geo-localization benchmarks and methods predominantly target near-nadir scenarios, offering limited diversity in viewing angles and lacking the complexity needed to evaluate robustness under extreme perspective variations~\cite{zhu2023simple,deuser2023sample4geo,DAC,wu2024camp,li2024unleashing,chen2025mean,li2025robust}. As a result, current models often fail to generalize when applied to large off-nadir images, which are increasingly common in real-world UAV deployments. As shown in Figure~\ref{fig:dataset}, we evaluate the model from Game4Loc trained on GTA-UAV~\cite{ji2025game4loc}, which provides diverse environments but only near-nadir views. The performance drops sharply with increasing off-nadir angles, highlighting the limitations of current models trained on near-nadir images and the urgent need for benchmarks and methods that address significant viewpoint variations. To address this gap, we propose OffNadirLoc, a novel benchmark specifically designed for large off-nadir UAV-to-satellite geo-localization. OffNadirLoc captures diverse oblique viewing angles with significant geometric and appearance disparities, providing a challenging testbed for evaluating cross-view localization under realistic conditions.

To effectively address the challenges posed by off-nadir matching, we propose a novel framework named ONLoc, which integrates structural scene understanding with multi-view relational learning. First, we introduce a structure-aware contextual weighting mechanism, which dynamically attenuates the influence of noisy or occluded regions, enabling the model to focus on geometrically reliable cues during feature aggregation. Second, we design a view-coherent learning strategy that leverages the natural grouping of multiple UAV views and their corresponding satellite image at each location. This strategy jointly enforces intra-location consistency and inter-location discriminability, moving beyond traditional pairwise contrastive learning to improve robustness under large viewpoint and appearance variations.

Extensive experiments conducted on four existing datasets, as well as the OffNadirLoc benchmark, demonstrate that our approach substantially outperforms state-of-the-art methods, validating the effectiveness of integrating structure-aware adaptation and multi-view supervision for robust large off-nadir geo-localization. We believe OffNadirLoc and our proposed framework will promote future research in practical cross-view geo-localization under challenging real-world conditions. Our main contributions are summarized as follows:

\begin{itemize}
     \item We construct OffNadirLoc, the first UAV-to-satellite geo-localization benchmark focused on large off-nadir views.
     
    \item We introduce a structure-aware contextual weighting mechanism that dynamically emphasizes geometrically salient regions while mitigating the effects of occlusions and geometric distortions.
    
    \item We design a view-coherent learning strategy that exploits multi-view UAV data per group to learn viewpoint-invariant and discriminative representations.
    
    \item Extensive experiments on OffNadirLoc and other benchmarks demonstrate that our approach consistently outperforms state-of-the-art methods, validating the superiority of our framework.

\end{itemize}

%% file: sec/2_related_work.tex
\section{Related Work}

\subsection{Datasets for Cross-View Geo-Localization}

Early cross-view geo-localization (CVGL) datasets primarily focused on matching street-view and \textit{satellite-view} imagery~\cite{lin2013cross,cheng2018crowd,liu2019lending,toker2021coming}. The University-1652~\cite{zheng2020university} was the first to additionally incorporate drone-view images, extending CVGL to multi-altitude aerial observations. It collected data from 1,402 target buildings across 1,652 universities worldwide, simulating UAV perspectives using oblique imagery captured from Google Earth. SUES-200~\cite{zhu2023sues} dropped street-view and used real UAV photos for 200 sites, systematically varying altitude (150–300 m) and weather. DenseUAV~\cite{denseuav} further increased difficulty by densely sampling UAV trajectories, yielding high inter-frame overlap and challenging retrieval. More recently, GTA-UAV~\cite{ji2025game4loc} built a contiguous-area UAV–satellite dataset from GTA-V, defining positives by the spatial IoU between UAV images and satellite patches to accommodate partial matches. Existing datasets mainly assume near-nadir views~\cite{uav}, offering little coverage of large off-nadir. We propose OffNadirLoc to fill this gap, with diverse viewing angles, target-focused evaluation, and diverse scenarios for robust cross-view testing.

\subsection{Methods for Cross-View Geo-Localization}

Early CVGL methods primarily relied on handcrafted descriptors or template matching~\cite{shan2015google}, which were sensitive to viewpoint and illumination changes. With the rise of deep learning, neural approaches such as MCVPlaces~\cite{workman2015wide} began to replace traditional pipelines. Subsequent methods, including CVM-Net~\cite{hu2018cvm} and DSM~\cite{shi2020looking}, improved cross-view correspondence via feature aggregation ~\cite{arandjelovic2016netvlad,shi2019spatial} and adaptive similarity modeling. The LCM model~\cite{ding2020practical} casts UAV geo-localization as a classification task by mapping image features to discrete location labels. In parallel, contrastive learning became dominant, learning viewpoint-invariant embeddings by pulling positive UAV–satellite pairs closer; for example, Sample4Geo~\cite{deuser2023sample4geo} employs InfoNCE~\cite{oord2018infoNCE} with hard-negative mining. ConGeo~\cite{mi2024congeo} is a model-agnostic contrastive learning framework that jointly applies single-view and cross-view objectives to align ground-view variants with aerial images. More recent advances use transformer-based models~\cite{dai2021transformer,yang2021cross,huang2022learning,zhu2022transgeo,zhang2024aligning} to aggregate multi-scale semantics for stronger spatial reasoning. DAC~\cite{DAC} introduces domain-alignment modules to reduce modality discrepancy. CAMP~\cite{wu2024camp} adds a position-aware local branch to capture fine-grained spatial cues. In summary, existing methods pay insufficient attention to filtering out redundant regions and to learning view-consistent representations, which limits their practicality. We propose ONLoc to focus on semantically meaningful regions and ensure robustness across viewpoints.

%% file: sec/3_dataset.tex

\section{OffNadirLoc Dataset}

To facilitate research on geo-localization under large off-nadir views, we introduce \textbf{OffNadirLoc}, a new benchmark specifically designed to train and evaluate UAV-to-satellite geo-localization in challenging viewing conditions. 

\subsection{Dataset Construction}

We constructed the OffNadirLoc dataset by systematically collecting UAV and satellite image groups across 44 geographically diverse regions worldwide, each covering approximately $8\mathrm{km} \times 6\mathrm{km}$. For each region, we extracted a high-resolution (level-18, $0.5$m/pixel) satellite map to serve as the reference gallery. These regions were carefully selected to encompass a wide range of representative terrain types, including dense urban centers, mountainous areas, coastal zones, ports, and airports, ensuring extensive environmental diversity.


Within each region, we randomly sampled 30–100 non-overlapping locations, each measuring approximately $350\mathrm{m} \times 200 \mathrm{m}$. To ensure that multi-view UAV images correspond precisely to the same physical site, we utilized Google Earth’s 3D engine to render UAV images from multiple off-nadir viewpoints, simulating realistic side-looking flight conditions. The UAV images were rendered at five extreme viewing angles ($70^\circ, 75^\circ, 80^\circ, 82^\circ,$ and $85^\circ$), with varied azimuth angles and altitudes to capture diverse perspectives. The rendered images have a resolution of $1000 \times 600$ pixels with realistic shadows and texture effects. Corresponding high-resolution satellite images were extracted to serve as reference views, forming natural cross-view groups linking UAV and satellite imagery for each location.

For rigorous evaluation, we split the dataset at the region level into training and test sets, ensuring that geographic regions in the test set are completely unseen during training. This geographical partitioning prevents data leakage and facilitates a realistic assessment of model generalization to novel environments.
 
In total, OffNadirLoc contains 9,736 UAV images and 1,657 satellite images, exhibiting substantial viewpoint variation, geometric distortions, and occlusions that pose significant challenges for existing geo-localization methods under large off-nadir UAV operations.
 
\noindent\textbf{Localization Setup} OffNadirLoc simulates practical UAV localization by matching each UAV image against a dense set of overlapping satellite patches, sliced from mosaics across all test regions. This reflects practical settings where the UAV's location is unknown and must be inferred from a broad search space. Due to the slicing, a UAV image may only partially match multiple satellite patches, leading to inherently imperfect correspondences. A retrieval is correct if the selected patch lies within a spatial threshold of the ground-truth center of the UAV image's field-of-view. This setup reflects realistic localization scenarios with ambiguous spatial extents, requiring fine-grained spatial reasoning, which makes the task both challenging and practically meaningful.

\begin{table}[H]
  \centering
  {\begingroup 
  \renewcommand{\arraystretch}{1.4} 
  \setlength{\tabcolsep}{2.2pt} 
  \scriptsize
  \begin{tabular}{@{}lcccc@{}}
    \toprule
    & Viewpoint Range & Scene Type & Evaluation Protocol & Cross-domain \\
    \midrule
    University & $10^\circ\!\sim\!30^\circ$ & Campus & Perfect & $\times$ \\
    SUES-200   & $10^\circ\!\sim\!30^\circ$ & Urban  & Perfect & $\times$ \\
    DenseUAV   & $0^\circ$               & Urban  & Perfect & $\times$ \\
    GTA-UAV    & $0^\circ\!\sim\!10^\circ$ & Game World & Partial & $\checkmark$ \\
    
    \textbf{OffNadirLoc} & \makecell[c]{$70^\circ,75^\circ,80^\circ,$ \\ $82^\circ,85^\circ$} & 
    \makecell[c]{Urban/Coastal/ \\ Airport} & Partial & $\checkmark$ \\

    \bottomrule
  \end{tabular}
  \vspace{-0.2cm}
  \caption{Comparison of existing UAV–satellite geo-localization datasets.}
  \vspace{-0.2cm}
\label{tab:dataset_comparison}
  \endgroup}
\end{table} 
\subsection{Comparison with Existing Datasets} 

Existing cross-view geo-localization datasets such as University-1652, SUES-200, DenseUAV, and GTA-UAV predominantly consist of images captured from nadir or near-nadir viewpoints ($\phi \in [0^\circ, 10^\circ]$). Although these datasets have facilitated significant progress in cross-view matching, they have several critical limitations:
1) \textbf{Limited Viewpoint Diversity}. The narrow range of viewing angles leads to only minor geometric distortions, which underestimates the complexity of perspective reasoning required for real-world UAV deployments. 2) \textbf{Simplified Matching Tasks}. Most existing datasets formulate retrieval as perfect matching between a UAV image and a tightly aligned satellite patch, assuming near-complete spatial overlap correspondence. This overlooks partial matching in realistic searches, where a UAV view may overlap with multiple satellite patches within a large gallery. 3) \textbf{Restricted Environmental Coverage}. Existing datasets primarily focus on campus or urban scenes, lacking representations of diverse terrains such as mountains, ports, or coastal areas. 



As shown in Table~\ref{tab:dataset_comparison}, OffNadirLoc addresses these gaps and introduces several unique challenges:
\begin{itemize}
    \item \textbf{Extreme off-nadir angles} introduce strong geometric distortions and appearance variations.

    \item \textbf{Diverse satellite gallery} covering multiple terrain types (urban, port, airport, mountain, coastal) increases scene complexity and distractors.

    \item \textbf{Partial evaluation protocol} accounts for imperfect overlap between UAV images and satellite patches, requiring models to handle spatial ambiguity.

    \item \textbf{Strict train-test split at the region level} ensures no geographic overlap, testing true generalization to unseen environments.
\end{itemize}

These features collectively create a challenging benchmark that pushes the limits of current geo-localization models and significantly broadens their applicability to diverse UAV deployment scenarios.

    

%% file: sec/4_method.tex
\section{Methods}

\begin{figure*}[h]
  \centering
  \includegraphics[width=0.9\linewidth]{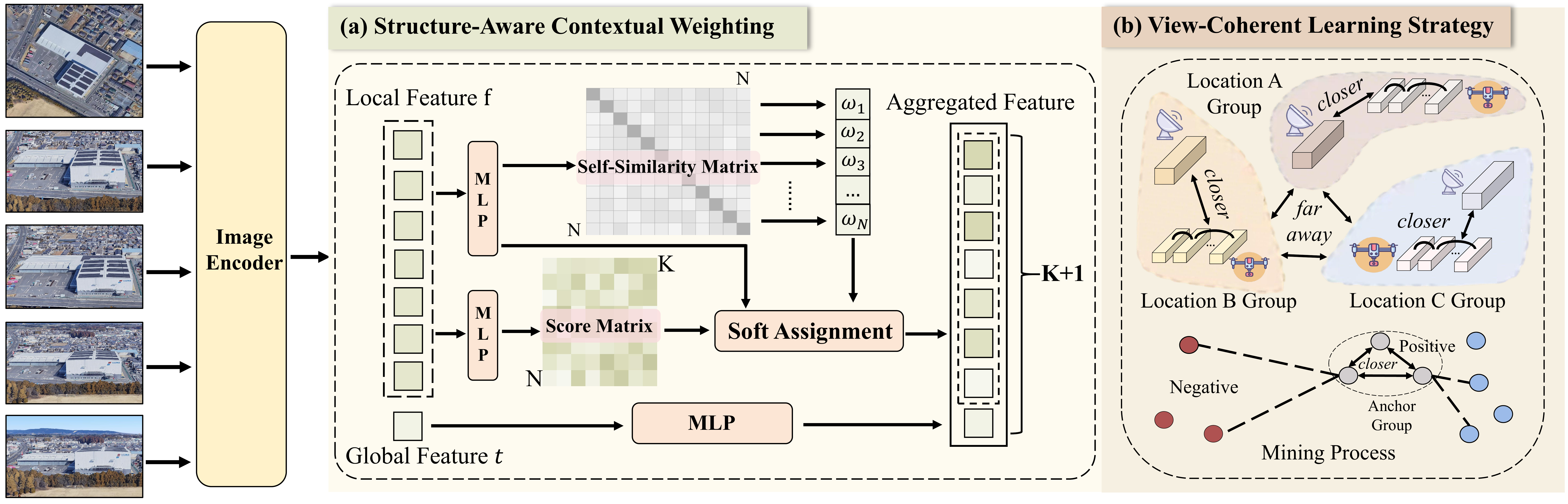} 
  \vspace{-0.2cm}
  \caption{Overview of the ONLoc pipeline. (a) Structure-Aware Contextual Weighting: Local and global features are extracted and aggregated using redundancy-aware clustering, generating robust structural representations under large off-nadir distortions.(b) View-Coherent Learning Strategy: Multi-view UAV and satellite images from the same location are grouped, and a group-wise similarity-based objective enforces cross-view and cross-modal consistency for viewpoint-invariant geo-localization.}
  \vspace{-0.2cm}
  \label{fig:pipeline}
\end{figure*}

As illustrated in Figure~\ref{fig:pipeline}, we propose a cross-view geo-localization framework, \textbf{ONLoc}, specifically designed to address the significant challenges posed by large off-nadir UAV-to-satellite geo-localization, where extreme viewpoint and appearance variations are present. Our method consists of two key components: (1) a Structure-Aware Contextual Weighting mechanism for robust representation learning under heavy geometric distortions, and (2) a View-Coherent Learning Strategy that leverages multi-view UAV observations per location to enhance cross-view discriminability and viewpoint invariance.

Given an input image $I$, we extract features using a ViT-based\cite{vit} visual encoder $\mathcal{E}(\cdot)$, which outputs a set of spatial local features $f = \{f_i\}_{i=1}^N$ with $f_i \in \mathbb{R}^l$, along with a global token $t \in \mathbb{R}^l$:
\begin{equation}
    f, t = \mathcal{E}(I)
\end{equation}
Here, $f \in \mathbb{R}^{l \times N}$ represents $N = H' \times W'$ local features each of dimension $l$, and $t$ is a global embedding summarizing the entire image.

Building on these features, ONLoc combines a structure-aware aggregation module and a multi-view training strategy, both designed to improve cross-view robustness and viewpoint invariance under extreme off-nadir conditions.

\subsection{Structure-Aware Contextual Weighting}
Off-nadir UAV images often exhibit extreme geometric distortion, making it difficult to rely on direct spatial correspondences. Features from repetitive regions, such as rooftops and roads, can dominate representations while offering limited geometric or semantic discriminability. Meanwhile, distinctive visual patterns like intersections and building contours are sparse and spatially inconsistent across views\cite{salad, burstvlad}. To address this, the SACW module adaptively suppresses redundant local features while preserving spatially unique and semantically meaningful structures via a redundancy-aware clustering mechanism.

To support structured aggregation, the local features \( f = \{f_i\}_{i=1}^N \) and the global token feature \( t \) are independently projected into a shared \( d \)-dimensional embedding space via two distinct two-layer MLPs \(\phi_{\text{l}}\) and \(\phi_{\text{g}}\), yielding \( f_i' = \phi_{\text{l}}(f_i) \) and \( t' = \phi_{\text{g}}(t) \). This projection reduces the feature dimension from \( l \) to \( d \) for computational efficiency and allows separate adaptation of local and global features for effective structural modeling and global context encoding.
 
All subsequent computations are performed on the projected features $f'$. To estimate spatial redundancy, we compute a pairwise similarity matrix across all spatial tokens:
\begin{equation}
S_{ij} = \langle \hat{f}_i', \hat{f}_j' \rangle,
\end{equation}
where $\hat{f}_i' = \frac{f_i'}{\|f_i'\|_2}$ denotes the $L_2$-normalized feature. The similarity matrix $S \in \mathbb{R}^{N \times N}$ captures self-similarity among spatial locations. Features that are similar to many others are likely redundant.

We then define a contextual redundancy score for each feature:
\begin{equation}
w_i = \sum_{j=1}^{N} \sigma(a S_{ij} + b),
\end{equation}
where $a, b$ are learnable scalars, and $\sigma(\cdot)$ is the sigmoid function. Larger $w_i$ values indicate higher redundancy.

To aggregate semantically meaningful structures, we introduce $K$ latent clusters, each serving as a semantic anchor. The soft assignment of each projected feature \( f_i' \) to cluster \( k \) is computed via an MLP head that outputs logits, followed by a normalized softmax:
\begin{equation}
p_{k,i} = \frac{\exp(\texttt{MLP}_{\text{score}}(f_i')_k)}{(w_i^p + \epsilon) \sum_{k'=1}^{K} \exp(\texttt{MLP}_{\text{score}}(f_i')_{k'})},
\end{equation}
where $p$ is a learnable exponent controlling redundancy suppression, and $\epsilon$ is a small constant for stability. This formulation encourages features from redundant regions to contribute less to the final representation.

Each cluster descriptor $c_k \in \mathbb{R}^d$ is obtained via weighted average of projected features:
\begin{equation}
c_k = \sum_{i=1}^{N} p_{k,i} \cdot f_i'.
\end{equation}

Finally, we form the image representation by concatenating the projected global token $t'$ with the $K$ cluster descriptors:
\begin{equation}
\mathbf{z} = \left[ t' \, \| \, c_1 \, \| \, \cdots \, \| \, c_K \right] \in \mathbb{R}^{(K+1) \times d}.
\end{equation}

This structured representation encodes both global semantics and spatially-discriminative structural cues, while adaptively discounting feature redundancy. It provides a robust basis for cross-view comparison under large viewpoint shifts.

\subsection{View-Coherent Learning Strategy}

Typical UAV-satellite geo-localization frameworks rely on pairwise contrastive loss~\cite{khosla2020supervised} between UAV and satellite image pairs. However, in large off-nadir scenarios, a single geographic location is often captured from multiple UAV viewpoints with significant appearance variations. Pairwise supervision alone cannot fully exploit these multi-view relationships to learn robust, viewpoint-invariant representations.
  
To leverage this multi-view information, we propose a view-coherent learning strategy that organizes supervision at the level of semantic groups rather than image pairs. Specifically, for each geographic location, we collect a set of $M$ UAV images captured from diverse off-nadir angles and a single satellite image. These $M+1$ images form a semantic group that represents the same physical location from multiple perspectives.

Assuming a training set comprising $N$ such groups, for the $i$-th group, we denote the satellite image as $\mathbf{I}^s_i$ and the UAV images as $\{\mathbf{I}^u_{i,1}, \dots, \mathbf{I}^u_{i,M}\}$. Using a feature extractor with adaptive weighting as described above, we obtain aggregated image embeddings $\mathbf{f}^s_i$ and $\{\mathbf{f}^u_{i,1}, \dots, \mathbf{f}^u_{i, M}\}$ for the satellite and UAV views respectively.

Rather than treating each UAV-satellite pair as an independent training signal, we enforce group-wise similarity constraints. Specifically, all features within the same group (regardless of modality or viewpoint) are treated as \emph{positives}, while features from different groups correspond to \emph{negatives}. This formulation enables the model to learn viewpoint-invariant representations that remain coherent within a location and discriminative across locations.

To optimize the group-wise similarity structure, we adopt the Multi-Similarity framework~\cite{multisimilarity}, which allows flexible and dense sampling of positive and negative pairs within each semantic group. For each embedding $\mathbf{f}$ (either UAV or satellite), we define its positive set $\mathcal{P}$ as all other embeddings from the same group (i.e., same geographic location), and its negative set $\mathcal{N}$ as embeddings from all other groups. 
The overall objective for a single anchor is formulated as:
\vspace{-0.1cm}
\begin{equation}
\mathcal{L}(\mathbf{f}) = \frac{1}{|\mathcal{P}|} \sum_{\mathbf{f}^+ \in \mathcal{P}} \ell_{p}(s(\mathbf{f}, \mathbf{f}^+)) + \frac{1}{|\mathcal{N}|} \sum_{\mathbf{f}^- \in \mathcal{N}} \ell_{n}(s(\mathbf{f}, \mathbf{f}^-)),
\end{equation}

where $s(\cdot, \cdot)$ denotes the cosine similarity, and $\ell_{p}$ and $\ell_{n}$ are sample-wise penalties that focus on mining informative positive and negative pairs. For instance, $\ell_{p}$ assigns higher weight to less similar positives, while $\ell_{n}$ emphasizes negatives with high similarity to the anchor.

\begin{figure*}[!h]
  \centering
  \includegraphics[width=1\linewidth]{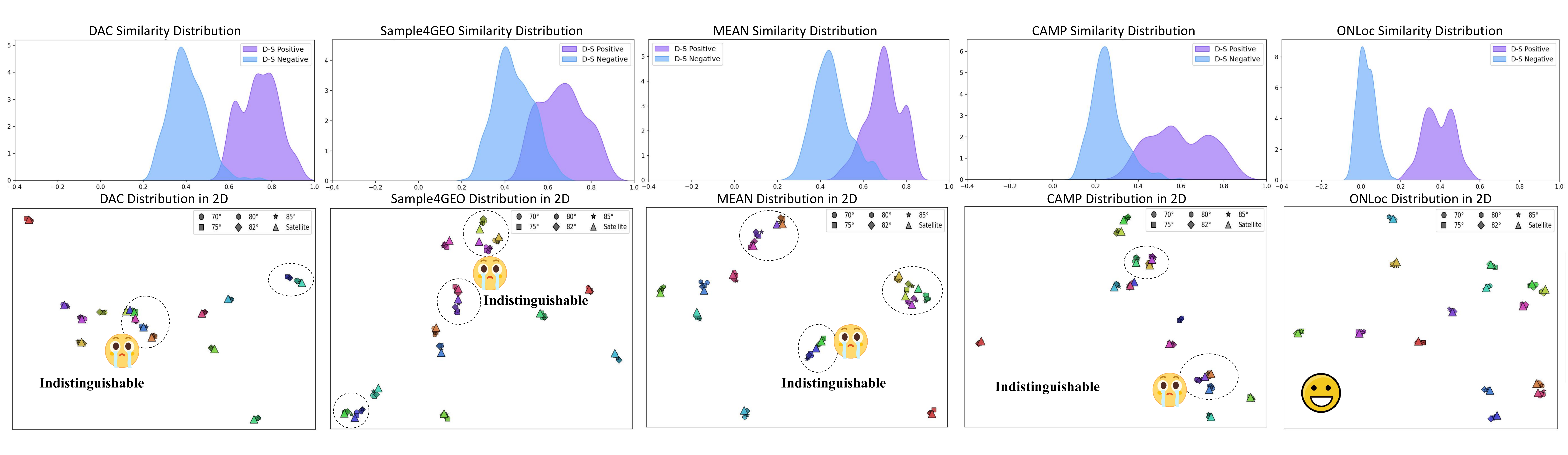} 
  \vspace{-0.8cm}
  \caption{The top row presents similarity distribution visualizations for different methods, while the bottom row shows feature embeddings projected into the 2D space. Triangles, circles, rectangles, hexagons, diamonds, and stars denote the satellite image and UAV images at 70°, 75°, 80°, 82°, and 85°, respectively, and different colors indicate different localizations. A total of 15 localization examples are included.}
  \vspace{-0.3cm}
  \label{fig:umap}
\end{figure*}

Unlike conventional contrastive learning schemes that rely on fixed anchor-positive assignments (e.g., satellite$\leftrightarrow$UAV), our formulation systematically mines all intra-group (cross-view, cross-modal) similarities, capturing the full relational structure among viewpoints. This allows the model to align all UAV views with the satellite representation and enforce cross-consistency among UAV images themselves, thus enabling the emergence of viewpoint-invariant, semantically consistent embeddings.

%% file: sec/5_experiment.tex

\section{Experiments}
\subsection{Implementation Details}

We adopt DINOv2-Base~\cite{oquab2023dinov2} as the image encoder, with approximately 86M parameters and a 14×14 patch size. We employ the Adam optimizer~\cite{loshchilov2017decoupled} with a decoupled weight decay of $9.5 \times 10^{-9}$. The initial learning rate is set to $6 \times 10^{-5}$ with a batch size of 32, and training runs for 20 epochs. During training, satellite image crops are generated from XML annotations, ensuring alignment with the corresponding UAV images. 
In evaluation, we partition the satellite maps—each approximately 8 km × 6 km—into partially overlapping 200 m × 200 m patches to construct a reference feature index. For each query feature, we retrieve the most similar patch from the reference library based on cosine similarity. Following VIGOR~\cite{zhu2021vigor}, a retrieval is considered correct if the patch comes from the same basemap and its IoU with the ground truth region exceeds 0.14. This patch-level evaluation reflects realistic deployment challenges in large-area UAV geo-localization. We report standard retrieval metrics: Recall@k ($k=1,3,5$) and Average Precision (AP)~\cite{zheng2020university,zhu2023sues,denseuav,zhang2023cross,ji2025game4loc}.

\subsection{Evaluation on the OffNadirLoc Benchmark}
We compare ONLoc with state-of-the-art (SoTA) methods~\cite{deuser2023sample4geo,DAC,wu2024camp,mi2024congeo,ji2025game4loc,chen2025mean} on the OffNadirLoc benchmark. 
Because prior works employ different backbones, we normalize this factor by using ConvNeXt and DINOv2 backbones for all methods. This controls for backbone-induced variance and isolates the contribution of model design, ensuring a fair comparison.
\begin{table}[h]
    \renewcommand{\arraystretch}{1}
    \centering
    \small
    \setlength{\tabcolsep}{6pt}
    \begin{tabular}{lcccc}
        \toprule
        \textbf{Method} & \textbf{R@1} & \textbf{R@3} & \textbf{R@5} & \textbf{AP} \\
        \midrule
        \multicolumn{5}{c}{\textbf{ConvNeXt Backbone}} \\
        \midrule
        Sample4Geo~\cite{deuser2023sample4geo}  & 46.80\% & 58.59\% & 65.32\% & 50.70\% \\
        DAC~\cite{DAC}         & 45.37\% & 59.01\% & 65.07\% & 49.99\% \\
        CAMP~\cite{wu2024camp}        & 47.98\% & 59.76\% & 65.24\% & 51.61\% \\
        MEAN~\cite{chen2025mean}        & 41.25\% & 57.32\% & 62.29\% & 46.95\% \\
        Game4Loc~\cite{ji2025game4loc}    & 38.64\% & 52.44\% & 59.34\% & 44.03\% \\
        ConGEO~\cite{mi2024congeo}      & 50.25\% & 62.46\% & 68.43\% & 54.01\% \\
        \textbf{ONLoc (Ours)} & \textbf{53.20\%} & \textbf{65.82\%} & \textbf{70.03\%} & \textbf{56.35\%} \\
        \midrule
        \multicolumn{5}{c}{\textbf{DINOv2 Backbone}} \\
        \midrule
        Sample4Geo~\cite{deuser2023sample4geo} & 65.74\% & 72.47\% & 77.95\% & 68.53\% \\
        DAC~\cite{DAC}         & 65.57\% & 77.86\% & 82.49\% & 68.59\% \\
        CAMP~\cite{wu2024camp}       & 63.80\% & 75.51\% & 80.72\% & 66.86\% \\
        MEAN~\cite{chen2025mean}       & 64.23\% & 76.26\% & 79.71\% & 66.22\% \\
        Game4Loc~\cite{ji2025game4loc}   & 64.90\% & 76.60\% & 80.56\% & 67.37\% \\
        ConGEO~\cite{mi2024congeo}     & 65.49\% & 78.87\% & 82.91\% & 68.49\% \\
        \textbf{ONLoc (Ours)} & \textbf{72.64\%} & \textbf{81.23\%} & \textbf{84.76\%} & \textbf{73.34\%} \\
        \bottomrule
    \end{tabular}
    \caption{Performance on OffNadirLoc benchmark with different backbone architectures. 
    The upper part shows results based on ConvNeXt backbone, while the lower part presents results based on DINOv2 backbone.}
    \vspace{-0.6cm}
    \label{tab:offnadir_results}
\end{table}

\begin{table*}[htbp]
    \centering
    \footnotesize
    \setlength{\tabcolsep}{5pt}
    \scalebox{0.9}{
    \begin{tabular}{l ccc ccc ccc ccc}
        \toprule
        & \multicolumn{3}{c}{\textbf{University-1652}} & \multicolumn{3}{c}{\textbf{SUES-200}} & \multicolumn{3}{c}{\textbf{DenseUAV}} & \multicolumn{3}{c}{\textbf{GTA-UAV}} \\
        \cmidrule(lr){2-4} \cmidrule(lr){5-7} \cmidrule(lr){8-10} \cmidrule(lr){11-13}
        \textbf{Method} & \textbf{R@1} & \textbf{R@5} & \textbf{AP} & \textbf{R@1} & \textbf{R@5} & \textbf{AP} & \textbf{R@1} & \textbf{R@5} & \textbf{AP} & \textbf{R@1} & \textbf{R@5} & \textbf{AP} \\
        
        \midrule
        \multicolumn{13}{c}{\textbf{Zero-shot transfer capability after training on OffNadirLoc}} \\
        \midrule
        
        Sample4Geo~\cite{deuser2023sample4geo}      & 72.35\% & 89.59\% & 76.23\% & 85.43\% & 96.86\% & 87.93\% & 16.73\% & 42.87\% & 13.09\% & 33.31\% & 56.74\% & 43.35\% \\
        DAC~\cite{DAC}             & 75.08\% & 91.49\% & 78.73\% & 86.63\% & 97.84\% & 89.02\% & 19.09\% & 45.77\% & 14.40\% & 33.63\% & 57.22\% & 43.89\% \\
        CAMP~\cite{wu2024camp}            & 72.67\% & 90.62\% & 76.66\% & 87.70\% & 97.62\% & 89.86\% & 17.76\% & 45.82\% & 13.96\% & 36.58\% & 60.14\% & 46.82\% \\
        MEAN~\cite{chen2025mean}             & 76.63\% & 92.35\% & 80.11\% & 90.38\% & \textbf{98.78\%} & 92.14\% & 21.06\% & 49.21\% & 15.68\% & 34.51\% & 56.95\% & 44.48\% \\
        Game4Loc~\cite{ji2025game4loc}           & 67.65\% & 86.54\% & 71.86\% & 82.38\% & 91.75\% & 83.13\% & 15.83\% & 40.63\% & 11.31\% & 16.82\% & 35.36\% & 25.07\% \\
        ConGEO~\cite{mi2024congeo}            & 79.78\% & 93.03\% & 82.76\% & 87.00\% & 98.18\% & 89.50\% & \textbf{22.31\%} & \textbf{55.73\%} & 17.80\% & 41.80\% & 63.74\% & 51.16\% \\
        \textbf{ONLoc(Ours)} & \textbf{79.87\%} & \textbf{93.23\%} & \textbf{82.90\%} & \textbf{91.38\%} & 97.68\% & \textbf{92.28\%} & 18.28\% & 49.29\% & \textbf{25.28\%} & \textbf{56.82\%} & \textbf{75.90\%} & \textbf{62.06\%} \\
        
        \midrule
        \multicolumn{13}{c}{\textbf{Comparison after training on near-nadir datasets}} \\
        \midrule
        
        Sample4Geo~\cite{deuser2023sample4geo}      & 93.78\% & 97.79\% & 94.71\% & 97.60\% & 99.74\% & 98.08\% & 61.09\% & 89.62\% & 53.73\% & 44.53\% & 68.49\% & 55.02\% \\
        DAC~\cite{DAC}              & 94.66\% & 98.38\% & 95.53\% & 97.48\% & 99.34\% & 97.90\% & 76.53\% & 95.45\% & 72.50\% & 45.98\% & 75.11\% & 58.11\% \\
        CAMP~\cite{wu2024camp}             & 92.84\% & 98.52\% & 93.98\% & 97.25\% & 99.60\% & 97.84\% & 82.93\% & 97.17\% & 76.82\% & 44.44\% & 73.53\% & 56.64\% \\
        MEAN~\cite{chen2025mean}             & 94.30\% & 98.58\% & 94.32\% & \textbf{98.56\%} & \textbf{99.93\%} & \textbf{98.88\%} & 80.01\% & 96.87\% & 73.64\% & 40.95\% & 65.81\% & 51.50\% \\
        Game4Loc~\cite{ji2025game4loc}           & 91.88\% & 97.38\% & 93.17\% & 97.04\% & 99.12\% & 97.69\% & 83.05\% & 96.22\% & 76.80\% & 57.13\% & 81.09\% & 65.90\% \\
        ConGEO~\cite{mi2024congeo}             & 92.98\% & 97.66\% & 94.06\% & 97.13\% & 99.89\% & 97.75\% & 85.97\% & 97.34\% & 76.49\% & 63.21\% & 83.04\% & 67.08\% \\
        \textbf{ONLoc(Ours)} & \textbf{95.65\%} & \textbf{99.03\%} & \textbf{96.45\%} & 97.99\% & 99.40\% & 98.24\% & \textbf{88.72\%} & \textbf{98.20\%} & \textbf{80.86\%} & \textbf{65.29\%} & \textbf{83.25\%} & \textbf{67.16\%} \\
        \bottomrule
    \end{tabular}
    }
        \caption{Comprehensive performance comparison for the drone-to-satellite task on near-nadir datasets. The top part shows the zero-shot transfer capability, while the bottom part shows the performance after training on these datasets. The best results are shown in bold.}
        \vspace{-0.2cm}
    \label{tab:combined-nearnadir-results}
\end{table*}

\begin{figure}[h]
  \centering
  \includegraphics[width=1\linewidth]{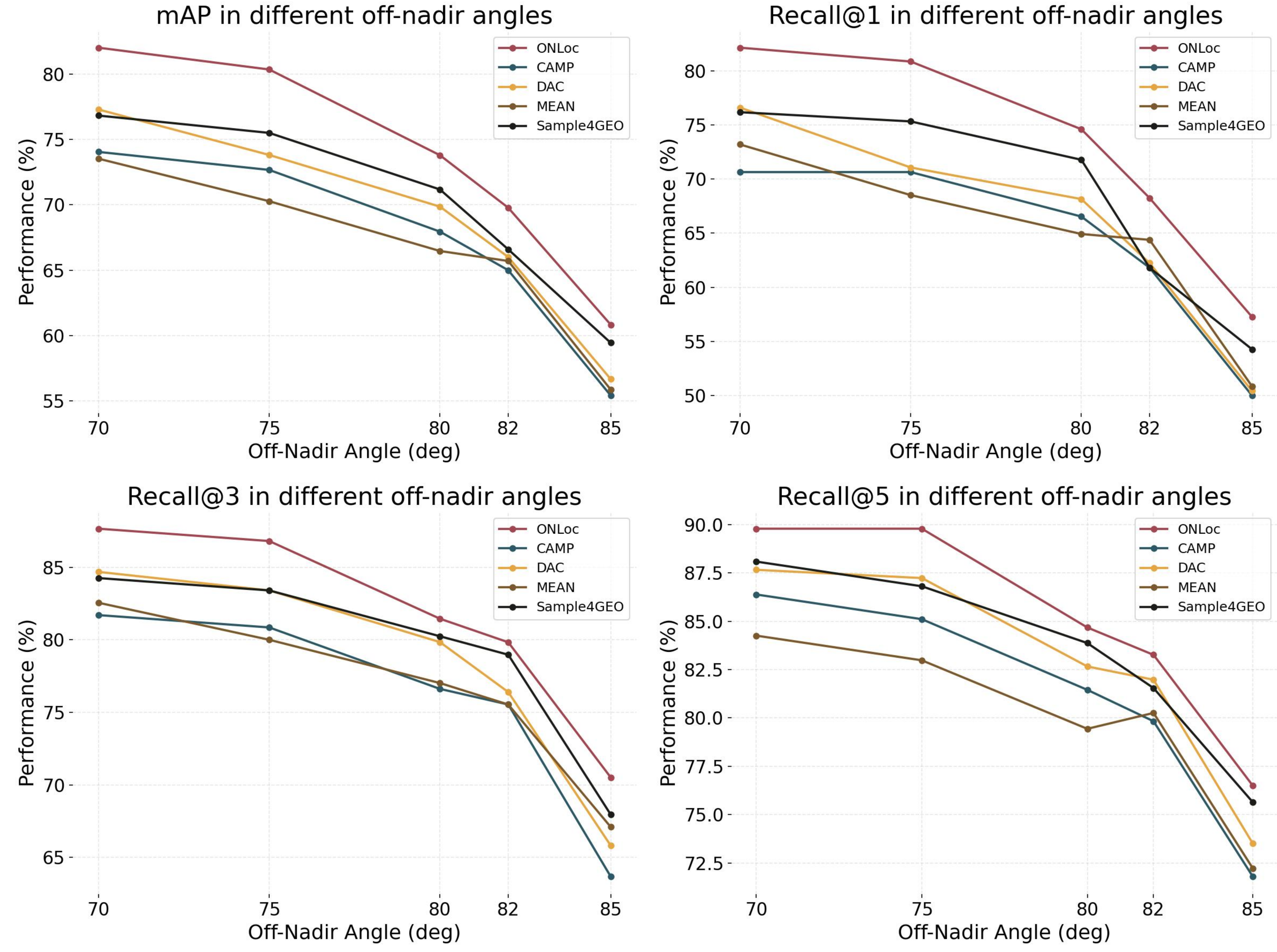} 
  \caption{Performance on the OffNadirLoc benchmark under large off-nadir UAV views (70°,75°,80°,82°,85°). Four panels report Recall@1, Recall@3, Recall@5, and mAP for five methods (ONLoc, CAMP, DAC, MEAN, Sample4GEO). } 
  \label{fig:linechart}
  \vspace{-0.2cm}
\end{figure}

As shown in Table~\ref{tab:offnadir_results}, ONLoc achieves SoTA performance in both ConvNeXt and DINOv2 backbones. DINOv2 consistently provides stronger representations than ConvNeXt across all methods. Under matched settings, ONLoc remains the top performer, surpassing the second-best approach by 6.9\% in Recall@1 and 4.75\% in AP, indicating superior robustness under large off-nadir views. To further analyze the feature behavior, we conduct two visualization studies. First, as shown in Figure~\ref{fig:umap}, we compare methods via similarity distributions and 2D embeddings: the top row shows that ONLoc produces a clearer bimodal similarity distribution, reflecting better separability between positives and negatives; the bottom row reveals that UAV views from the same location but different pitch angles form tighter, more compact clusters with larger inter-cluster margins. Second, as shown in Figure~\ref{fig:linechart}, we plot performance as a function of off-nadir angle (70° to 85°). ONLoc consistently outperforms prior methods across all angles in both Recall@k and mAP, demonstrating stable adaptability to extreme viewpoint changes.



These gains stem from two key factors. First, the structure-aware contextual weighting module suppresses the redundant and background-dominated regions prevalent in off-nadir UAV imagery, amplifying geometrically reliable and discriminative cues. This aspect is often overlooked by prior work, which primarily focuses on loss design while insufficiently addressing feature redundancy. Second, the view-coherent learning strategy treats all UAV views of a location and the corresponding satellite image as a unified semantic group, explicitly enforcing intra-location, cross-view, and cross-modal consistency while enhancing inter-location discrimination. In contrast to pairwise training that models each UAV–satellite pair independently, our grouping formulation leverages the intrinsic multi-view coherence within each site. Together, these components provide complementary strengths—spatial structural discrimination and viewpoint invariance—yielding robust cross-view localization in challenging large off-nadir scenarios.

\subsection{Zero-Shot Transfer Capability}

To evaluate the domain generalization ability of ONLoc, we conduct zero-shot transfer experiments where all models are trained solely on our OffNadirLoc dataset and directly tested on four widely used near-nadir benchmarks—University-1652~\cite{zheng2020university}, SUES-200~\cite{zhu2023sues}, DenseUAV~\cite{denseuav}, and GTA-UAV~\cite{ji2025game4loc}—without any fine-tuning. For SUES-200, results are averaged across four altitudes (150 m, 200 m, 250 m, 300 m). For GTA-UAV, we report results only on the cross-area subset, which better reflects real-world deployment scenarios. All baselines use the DINOv2 backbone for a fair comparison. The data in the upper portion of Table~\ref{tab:combined-nearnadir-results} indicates that our method achieves the best results in most cases across datasets and metrics. In the remaining cases, our method is still competitive, confirming its strong robustness and viewpoint invariance.
Performance on DenseUAV remains challenging for all methods, as the satellite gallery contains many densely overlapping patches differing by only a few pixels. Although several patches could be valid matches, the evaluation protocol accepts only the perfectly aligned one, making the distinction particularly difficult. Despite this, ONLoc maintains a clear advantage, demonstrating that training under extreme off-nadir conditions enables the model to capture semantically rich, structure-aware representations that transfer effectively to conventional UAV–satellite scenarios.

\subsection{Evaluation on Near-Nadir Datasets}

To further validate the versatility of our framework, we perform standard supervised training and evaluation on the same four near-nadir benchmarks with DINOv2 backbone. All competing methods are trained and tested within each dataset under identical settings to ensure fairness. The results in the lower portion of Table~\ref{tab:combined-nearnadir-results} show that ONLoc achieves highly competitive results across all benchmarks.
The improvement is particularly evident on datasets with complex scenes, such as DenseUAV and GTA-UAV.
Although ONLoc is primarily designed for off-nadir scenarios, the combination of structure-aware aggregation and view-coherent learning allows it to balance global semantics with spatially consistent local cues—capabilities that remain advantageous in near-nadir environments as well.
These findings confirm that the robust representations learned under off-nadir conditions generalize effectively across different domains and UAV flight geometries.

\subsection{Ablation Studies}
 We conduct component ablations on the OffNadirLoc dataset to evaluate the individual and joint contributions of the proposed modules. Results are summarized in Table~\ref{tab:ablation_studies}. 

\begin{table}[h]
    \centering
    \small
    \renewcommand{\arraystretch}{1} 
    \setlength{\tabcolsep}{3pt}
    \begin{tabular}{lcccc}
        \toprule
        \textbf{Setting / Component} & \textbf{R@1} & \textbf{R@3} & \textbf{R@5} & \textbf{AP} \\
        \midrule
        Baseline model & 62.29\% & 74.66\% & 79.12\% & 65.63\% \\
        + VCLS  & 66.58\% & 76.85\% & 80.89\% & 69.01\% \\
        + SACW  & 70.71\% & 79.97\% & 83.25\% & 71.72\% \\
        \textbf{Full model} & \textbf{72.64\%} & \textbf{81.23\%} & \textbf{84.76\%} & \textbf{73.34\%} \\
        \midrule
    \end{tabular}
    \vspace{-0.2cm}
    \caption{Component ablation on our proposed method. The first row corresponds to the DINOv2 baseline model without any improvement. The following two lines demonstrate the performance of the model with only one component added. The last row combines the two components.}
    \vspace{-0.1cm}
    \label{tab:ablation_studies}
\end{table}

\noindent\textbf{Effect of View-Coherent Learning Strategy.} 
We use “DINOv2 backbone + global token + Multi-Similarity loss” as the baseline and then introduce VCLS while keeping all other factors unchanged. Unlike the baseline, which treats multi-view UAV images of the same location as independent samples, VCLS assembles, within each mini-batch, M UAV images and their corresponding satellite image into a semantically coherent group. Within grouping, the constraint explicitly pulls features across views and modalities of the same location while pushing apart features from different locations, thereby enforcing intra-location coherence and inter-location separability. This group-level supervision exploits multi-view structure beyond fixed pairwise alignments, reducing representation drift under large off-nadir appearance changes and occlusions. On OffNadirLoc, adding VCLS alone yields consistent gains: Recall@1 improves from 62.29\% to 66.58\% and AP from 65.63\% to 69.01\%, indicating that group-consistent supervision better matches the intrinsic demands of large off-nadir scenarios and yields more viewpoint-robust semantics.

\noindent\textbf{Effect of Structure-Aware Contextual Weighting.}  
We next assess SACW in isolation (without VCLS) to quantify the benefit of structure-aware aggregation under severe geometric distortion. SACW computes redundancy-aware similarities among local tokens, performs soft clustering with a learnable redundancy suppression factor to down-weight highly self-similar regions, and preserves discriminative structural cues such as intersections and building contours. Under identical optimization and data settings, introducing SACW leads to improvements over the baseline: Recall@1 increases from 62.29\% to 70.71\% and AP from 65.63\% to 71.72\%. This highlights that in large off-nadir conditions, prioritizing “what to aggregate” is critical—SACW substantially mitigates the influence of repetitive textures and occlusion noise, steering cross-view matching toward geometrically reliable, discriminative regions.

\noindent\textbf{Effect of Combined Components.} 
When combined, SACW and VCLS are complementary: SACW enhances spatial discrimination and suppresses redundancy at the feature level, while VCLS enforces group-wise multi-view consistency at the supervision level. Together, they improve over the baseline by 10.35\% in Recall@1 and 7.71\% in AP, respectively, demonstrating a coherent end-to-end benefit from “feature selection” to “group supervision”.

\begin{table}[!h]
    \renewcommand{\arraystretch}{1}
    \centering
    \small
    \setlength{\tabcolsep}{4.5pt}
    \begin{tabular}{lcccc}
        \toprule
        \textbf{Hyperparameter} & \textbf{R@1} & \textbf{R@3} & \textbf{R@5} & \textbf{AP} \\
        \midrule
        cluster=32  & 69.36\% & 78.87\% & 81.40\% & 70.61\% \\
        cluster=48  & 69.95\% & 78.96\% & 82.15\% & 71.19\% \\
        \textbf{cluster=64}  & \textbf{72.64\%} & \textbf{81.23\%} & \textbf{84.76\%} & \textbf{73.34\%} \\
        cluster=80  & 69.78\% & 81.14\% & 84.34\% & 72.19\% \\
        \bottomrule
    \end{tabular}
    \vspace{-0.2cm}
    \caption{Ablation study on the hyperparameters of our proposed module. We keep the other hyperparameters consistent and only change the number of clusters in the SACW module. }
    \vspace{-0.2cm}
    \label{tab:ablation_hyperparams}
\end{table}
\noindent\textbf{Effect of Hyperparameters in the Proposed Module.}
We further analyze the key hyperparameters of our aggregation module—the number of clusters in the SACW module. As shown in Table~\ref{tab:ablation_hyperparams}, too few clusters (e.g., 32) limit the representational capacity of local feature aggregation, resulting in suboptimal performance. Conversely, overly large cluster numbers (e.g., 80) introduce redundant partitions and increase noise, harming discriminability. The best performance emerges at 64 clusters, which offers an effective balance between feature compactness and diversity.

%% file: sec/6_conclusion.tex
\section{Conclusion}
In this work, we tackle the underexplored problem of large off-nadir UAV-to-satellite geo-localization by introducing OffNadirLoc, a new benchmark with challenging oblique views. Our proposed ONLoc framework combines structure-aware weighting and view-coherent learning to address severe perspective distortions and multi-view consistency. Experiments show ONLoc outperforms state-of-the-art methods, highlighting its effectiveness. We hope this work fosters further research in robust cross-view geo-localization under real-world conditions.

\subsection*{Acknowledgement}
This work was supported in part by the National Natural Science Foundation of China (No.~62576279).